\documentclass{article}
\usepackage{aaai}
\usepackage{times}
\usepackage{helvet}
\usepackage{courier}
\usepackage{amsmath,amssymb}
\usepackage{graphicx}

\newtheorem{definition}{Definition}
\newtheorem{proposition}{Proposition}

\newenvironment{example}{{\vspace*{12pt}\noindent \bf Example} \ }{\vspace*{12pt}}

\title{Ordinal Conditional Functions for Nearly Counterfactual Revision}
\author{ Aaron Hunter\\
BCIT\\
Burnaby, BC, Canada\\
aaron\_hunter@bcit.ca
}

\nocopyright

\begin{document}

\maketitle

\begin{abstract}
We are interested in belief revision involving conditional statements where the antecedent is almost certainly false.  To represent such problems, we use Ordinal Conditional Functions that may take infinite values.  We model belief change in this context through simple arithmetical operations that allow us to capture the intuition that certain antecedents can not be validated by any number of observations.  We frame our approach as a form of finite belief improvement, and we propose a model of conditional belief revision in which only the ``right'' hypothetical levels of implausibility are revised.

\end{abstract}

\section{Introduction}
The theory of belief change is concerned with the way agents incorporate new information.  Typically, the focus is on new information that is given as a propositional formula.  In this paper, we are concerned with situations where an agent needs to revise by a conditional where the antecedent is almost certainly false.  More precisely, we consider antecedents that will not be believed given any finite amount of ``regular'' supporting evidence.  We represent the degree of belief in such formulas using Ordinal Conditional Functions that may take infinite values, and we provide an approach to conditional revision based on basic ordinal arithmetic.

This paper makes several contributions to existing work on belief change.\footnote{This paper contains results that have been published in \cite{Hunter15} and \cite{Hunter16}.}  First, we demonstrate that 
a simple algebra of belief change in the finite case extends naturally to the infinite case, giving a form of belief improvement.
In the process, we demonstrate that there are natural examples in commonsense reasoning where multiple levels of infinite implausibility are actually useful.  In particular, we introduce a natural approach to revision by conditional statements with little in the way of new formal machinery.  

\subsection{Motivating Example}
Consider the following claims:
\begin{enumerate}
\item $heavy$: Your dog is overweight.
\item $fly$: Your dog can fly.
\item $hollow | fly$:  If your dog can fly, then it has hollow bones.
\end{enumerate}
The first two claims are simple declarative statements.  But note that there is a clear difference in the amount of evidence needed to convince the agent to believe each claim.  For (1), it presumably takes some finite number of reports from a trusted source.  For (2), it seems unlikely that any finite number of reports would be convincing.  This statement is almost certainly false, though it is possible to imagine a situation that would convince an agent to believe it.

The third statement is a conditional with a highly unlikely antecedent.  Nevertheless, the perceived ``impossibility'' of (2) does not mean that (3) is free of content.  Revision by (3) should change an agent's beliefs in a counterfactual sense; they may need to change their beliefs about hollow bones in some hypothetical scenario.  
Moreover, if ever the notion of flying dogs becomes believable, then this report will take on significance at the level of factual beliefs.  In this paper, we refer to claims such as (3) as {\em nearly counterfactual}.  We will provide a formal characterization of such claims, as well as a suitable approach to revision.

\section{Preliminaries}
\subsection{Belief Revision}
{\em Belief revision} is the belief change that occurs when new information is presented to an agent
with some prior, possibly contradictory, set of beliefs.  We assume an underlying propositional signature ${\mathbf P}$.  An interpretation over ${\mathbf P}$ is called a {\em state}, while a logically closed set of formulas over ${\mathbf P}$ is called a {\em belief set}.  A belief revision operator is a function that combines the initial belief set and a formula to produce a new belief set.

Formal approaches to belief revision typically require an agent to have some form of {\em ordering} or {\em ranking} that gives the relative plausibility of possible states.  For example, in the well-known AGM approach, total pre-orders over states are used to represent the perceived likelihood of each state \cite{AlchourronGardenforsMakinson85,KatsunoMendelzon92}.  
Unfortunately, this approach does not handle the problem of {\em iterated belief revision}.  Related work has addressed
iterated revision by explicitly specifying how the ordering changes, rather than just the belief set \cite{DarwichePearl97,BoothMeyer06,JinThielscher07}.

\subsection{Ordinal Conditional Functions}
An ordinal conditional function (OCF) is a function that maps each state to an ordinal \cite{Spohn88,Williams94}.  In this approach, strength of belief is captured by ordinal precedence.  Hence, if $r$ is an OCF and $r(s) < r(t)$, then $s$ is a more plausible state than $t$.  There is an obvious advantage to this approach in that a ranking function is clearly more expressive than a total pre-oder.    

While the orginal definition allows the range of an OCF to be the class of all ordinals, in existing work it is common
to restrict the range to the natural numbers, possibly with an additional symbol $\infty$ representing impossibility.  In this paper, we will actually use a slightly larger range; so we need to briefly review ordinal arithmetic.\footnote{
It is beyond the scope of this paper to give a complete treatment of infinite ordinals, and ordinal arithmetic.  In the discussion here, we skip over fundamental set theory, and the fact that order-types are defined in terms of set-containment.  We refer the reader to \cite{Devlin93} for an excellent introduction.}
For our purposes, it is sufficient to note that ordinals are actually sets defined by an ``order type.''  The finite ordinals are the natural numbers.  The order type of the natural number $n$ is unique, because it is the only ordinal that has exactly $n-1$ preceding ordinals.  The first infinite ordinal is $\omega$, the set of all natural numbers.  Every countably infinite subset of the natural numbers is order-isomorphic to $\omega$.

It is easy to construct a countably infinite set that is not order isomorphic to $\omega$:  just add another symbol $\infty$ at the end that is larger than every natural number.  The ordinal that defines the order type of this set is written $\omega+1$.  Similarly, there exists a distinct ordinal $\omega +n$ for any natural number $n$.  And if we add a complete copy of the natural numbers, then we have the ordinal $\omega+\omega$ which is normally written as $\omega \cdot 2$.  We can procede in this manner indefinitely to define a countably infinite sequence of ordinals.  By taking powers, we can get even more order types; we will not delve further into this topic. 

Ordinal addition can be understood in terms of the informal discussion above.  Given ordinals $\alpha$ and $\beta$, the ordinal $\alpha+\beta$ has the order type obtained by taking a set with order type $\alpha$ and then appending a set with order type $\beta$ where all the elements of $\beta$ follow the elements of $\alpha$ in the underlying ordering.  For finite ordinals, this coincides with the usual notion of addition.  For infinite valued ordinals it does not.  Note for example that $1+\omega = \omega$; adding a number that precedes $0$ does not change the order type, because the resulting structure is isomorphic to the natural numbers.  On the other hand $\omega+1\ne\omega$.  So ordinal addition is not commutative.  It is also worth noting that ordinal subtraction is, in general, not well defined.  In particular, it is not possible to define subtraction by $\omega$.

\section{Belief Change as Ordinal Arithmetic}
Although our goal is to address revision by conditionals, we first introduce a simple approach to belief change based on the addition of ordinals.  This will allow us to precisely define the notion of a nearly counterfactual statement, which is important for the class of conditionals that we wish to consider.

\subsection{Restricted Domains}
The following definition allows us to define conditional functions over any set of ordinals.
\begin{definition} 
Let $S$ be a non-empty set of states and let $\Gamma$ be a collection of ordinals.  A $\Gamma$-CF ( $\Gamma$ conditional function) over $S$ is a function $r:S\rightarrow\Gamma$ such that $r(s)=0$ for some state $s$.
\end{definition}  
Note that the definition of $\Gamma$-CFs does not actually specify that $\Gamma$ is a {\em set}, because we do not wish to specify the underlying set theory in detail.  

\begin{figure}
    \centering
    \includegraphics[width=3in]{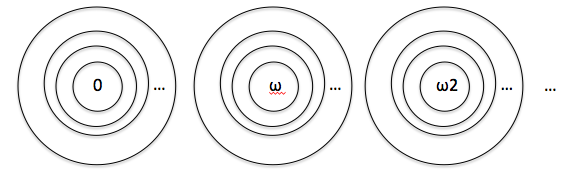}
    \caption{Visualizing $\omega^2$}
\end{figure}

Several special cases are immediate:
\begin{itemize}
\item Spohn's ordinal conditional functions are $\Omega$-CFs, where $\Omega$ is the collection of all ordinals.
\item The class of $\omega$-CFs coincides with the finite valued ranking functions common in the literature. 
\item The class of $(\omega+1)$-CFs is the set of ranking functions that can take finite values, as well as the single ``impossible'' plausiblity value $\infty$.  This is essentially equivalent to the possibilistic logic framework of \cite{DuboisPrade04}, that uses the ``necessity measure'' of 0.
\end{itemize}
In this paper, we are primarily interested in the class of $\omega^2$-CFs.  Note that $\omega^2$ can be specified as follows:  
$$\omega^2 = \bigcup\{\omega\cdot k + c \mid k,c\in\omega\}.$$
Hence, every element of $\omega^2$ can be written as $\omega\cdot k + c$ for some $k$ and $c$.  We think of these conditional functions as having countably many infinite levels of implausibility.  A picture of $\omega^2$ is shown in Figure 1.

If $r$ is a $\Gamma$-CF, we write 
$$Bel(r) = \{x \mid r(x)=0\}.$$
The {\em degree of strength} of a conditional function $r$ is the least $n$ such that $n=r(v)$ for some $v\not\in Bel(r)$.  Hence, the degree of strength is a measure of how difficult it would be for an agent to abandon the currently believed set of states.

\subsection{Finite Arithmetic on Conditional Functions}
In the finite case, belief change can be captured through addition on ranking functions.  Some variant of the following definition has appeared previously in published work by several authors; it is restated here and translated to our terminology.
\begin{definition}\label{def:rev}
Let $r_1$ and $r_2$ be $\omega$-CFs over $S$, and let $m$ be the minimum value of $r_1+r_2$.  Then $r_1\bar{+}r_2$ is the function on $S$ defined as follows:
$$r_1 \bar{+} r_2(x) = r_1(x)+r_2(x)-m.$$
\end{definition}
It is easy to check that this operation is associative, commutative, and that every element is invertible in the sense that, for each $r$ there is an $r^{\prime}$ such that $r\bar{+}r^{\prime}=0$.  Therefore, in terms of algebra, we say that the class of $\omega$-CFs is an {\em abelian group} under $\bar{+}$.

Note that Spohn's {\em conditionalization} can be seen as a special case of this algebra on ranking functions.  Let $r_1$ be a finite plausibility function representing the initial beliefs of an agent.  Let $\phi$ be a formula, let $d$ be a positive integer, and let $r_2$ be the ranking function defined as follows:
\[
  r_2(s) = \left\{
                    \begin{array}{ll}
                           0 \mbox{ if } s\models\phi \\
                           d \mbox{ otherwise }
                    \end{array}
                    \right.\\
\]
Then $r_1 \bar{+} r_2$ is equivalent to Spohn's conditionalization of $r_1$ by $\phi$ with strength $d$.  Similarly, if $r_2$ takes only two values and the degree of strength of $r_2$ is strictly larger than the degree of strength of $r_1$, then $r_1 \bar{+}r_2$ is AGM revision.

This approach does not extend to larger classes of ordinals.
\begin{proposition}
Let $\beta$ be an ordinal such that $\omega\in\beta$.  Then $\bar{+}$ is not well-defined over the class of $\beta$-CFs.
\end{proposition}
The problem is that subtraction is not defined for all pairs of (infinite) ordinals.

\begin{example}
Consider the motivating example.  We can define the following $(\omega+1)-CFs$:
\[
    r_1(s) = \left\{
                    \begin{array}{ll}
                           0 \mbox{ if } s\models\{fly\} \\
                           \omega \mbox{ otherwise }
                    \end{array}
                    \right.\\
\]
\[
   r_2(s) = \left\{
                    \begin{array}{ll}
                           \omega \mbox{ if } s\models\{fly\} \\
                           0 \mbox{ otherwise. }
                    \end{array}
                    \right.\\
\]
Normalized addition of $r_1$ and $r_2$  requires us to calculate $\omega - \omega$.  But this subtraction is not defined, so the calculation can not be completed.
\end{example}

This problem could be avoided by removing the normalization, but the result would no longer be an OCF.  If we want to work with ranking functions that are closed under some form of addition, then we must either modify the definition, or we must relax the constraint that the pre-image of 0 is non-empty.  We opt for the former.

\subsection{Finite Zeroing}
We define an algebra over $\omega^2$-CFs based on {\em finite zeroing}.  
The following relation will be useful in proving results.  In the definition, and in some future results, it is useful to
consider functions over ordinals that do not necessarily take 
the value 0 for any argument.  We use the general term $\Gamma$ ranking to refer to an arbitrary function from $S$
to $\Gamma$.\footnote{Konieczny refers to this kind of OCF as a {\em free OCF.}\cite{Konieczny09}}

\begin{definition}
For $\Gamma$ rankings $r_1$ and $r_2$, we write $r_1\sim r_2$ just in case the following condition holds for every pair of states $s, t$
$$r_1(s) < r_ 1(t) \iff  r_2(s) < r_ 2(t).$$
\end{definition}
Clearly, 
$\sim$ is an equivalence relation.

The intuition behind finite zeroing is that each conditional function can be categorized by its minimum value, in a manner that is useful for revision.   Given any $\omega^2$ ranking $r$, let $min(r)$ denote the minimum value $r(s)$.  Note that a minimum is guaranteed by the fact that the ordinals are well-ordered.
\begin{definition}
Let $r$ be an $\omega^2$ ranking with $\min(r)=\omega\cdot k + c$.  Then $k$ is the {\em degree} of $r$ and $c$ is the {\em finite shift}, written $deg(r)$ and $fin(r)$ respectively.
\end{definition}
We can use the degree and the finite shift to define the following operation.
\begin{definition}
Let $r$ be an $\omega^2$ ranking with $deg(r)=k$ and $fin(r)=c$.  Define $\bar{r}$ as follows.  Let $s$ be a state with $r(s)=\omega\cdot m + p$.
\begin{enumerate}
\item  If $m>k$, then $\bar{r}(s) = \omega\cdot(m-k)+c$.
\item If $m=k$, then $\bar{r}(s) = (p-c)$.
\end{enumerate}
\end{definition}
We call $\bar{r}$ the {\em finite zeroing} of $r$.  Intuitively, elements at the 
``lowest level'' are normalized to zero and elements at higher levels are shifted down by the degree of $r$. The following result is easy to prove.
\begin{proposition}
If $r$ is an $\omega^2$ ranking, then $\bar{r}$ is a $\omega^2$-CF and $r\sim\bar{r}$.
\end{proposition}
Hence, the finite zeroing of any ranking is an equivalent $\omega^2$-CF.
We can now extend the definition of $*$ to $\omega^2$-CFs.  
\begin{definition}
Let $r_1,r_2$ be $\omega^2$-CFs.  Then 
$$r_1*r_2 = \overline{r_1+r_2}.$$
\end{definition}
Using this definition, $*$ is consistent with $\bar{+}$ for $\omega$-CFs.  Hence, $*$ can capture standard belief revision operators (e.g., AGM, DP) by restricting to finite values and setting the degree of strength
of each function appropriately.  This is the natural extension of revision, therefore, to the case that allows infinite plausibility values.

\begin{example}
The motivating example over $\{heavy,fly\}$ can be captured by the following function:
\[
    r(s) = \left\{
                    \begin{array}{ll}
                           \omega \mbox{ if } s\models fly\\
			10\mbox{ if } s\models heavy\wedge\neg fly\\
                           0 \mbox{ otherwise }
                    \end{array}
                    \right.\\
\]
We let $*^n$ to denote a finite iteration of the $*$ operator. 
Suppose that, for each $V\in\{heavy,fly\}$, $r_V$ is an OCF such that $r_V(s)=2$ if and only if
$s\not\models V$.  The following are immediate:
\begin{itemize}
\item $r*^n r_{heavy}(s)=0$ iff $n\ge 5$.
\item $r*^n r_{fly}(s)\ne0$ for any $n$.
\end{itemize}
Hence, it takes 5 reports to convince the owner that their dog is overweight.  No finite number of reports will convince them that the dog can fly. 
\end{example}

In the $\omega^2$ case, the algebra obtained is not identical to the finite case.
\begin{proposition}
The class of $\omega^2$-CFs is a non-abelian group under $*$.  (i.e. it is closed, associative, and every element has an inverse, but it is not commutative). 
\end{proposition}
The fact that $*$ is not commutative has interesting consequences, as illustrated in the following example.

\begin{example}
Assume again that the vocabulary contains the predicates $\{heavy, fly\}$.  Define 
\[
   r_1(s) = \left\{
                    \begin{array}{ll}
                           \omega \mbox{ if } s\models fly \\
                           0 \mbox{ otherwise }
                    \end{array}
                    \right.\\
\]
\[
   r_2(s) = \left\{
                    \begin{array}{ll}
                           0 \mbox{ if } s\models\neg heavy \wedge fly \\
                           1 \mbox{ if } s\models heavy \wedge fly\\
			2   \mbox{ otherwise. }
                    \end{array}
                    \right.\\
\]
Hence, $r_1$ says that an agent believes dogs can not fly; moreover the agent essentially believes that a flying dog is an impossibility.  On the other hand, $r_2$ says that an agent believes that light dogs can fly - although the the strength of belief in this claim is only finite.  Moreover, $r_2$ gives an ordering over less plausible states as well.
Note that both $r_1$ and $r_2$ can be either an initial belief state or an observation.  The following calculations are immediate.
\[
   r_1*r_2(s) = \left\{
                    \begin{array}{ll}
                           \omega \mbox{ if } s\models\neg heavy \wedge fly \\
                           \omega + 1 \mbox{ if } s\models heavy \wedge fly\\
			0   \mbox{ otherwise. }
                    \end{array}
                    \right.\\
\]
\[
   r_2*r_1(s) = \left\{
                    \begin{array}{ll}
                           \omega \mbox{ if } s=\{fly\} \\
                           0 \mbox{ otherwise. }
                    \end{array}
                    \right.\\
\]

\end{example}

What is the significance of this example?  It shows that conditional beliefs from an observation can be maintained at higher plausibility levels.  In both cases, the underlying agent will not believe dogs can fly following revision.  But the first revision allows the ordering of states to be refined somewhat at the conditional level.  The second revision, on the other hand, washes away the finite level distinctions in the original belief set.  This is similar to AGM revision in the sense that recent information seems to carry some particular weight.  However, the infinite jumps in plausibility outweigh the preference for recency.


 
\section{Nearly Counterfactual Reasoning}
\subsection{Motivation}
In this section, we demonstrate how infinite-valued ordinal conditional functions can be useful for reasoning about conditional statements.  

\begin{example}
We return to the flying-dog example.  Suppose that we initially believe $\neg fly$ and $\neg hollow$; in other words, we believe that dogs do not fly and that dogs do not have hollow bones.   Now suppose we are told that flying dogs have hollow bones.  Informaly, we want to revise by the conditional statement $(hollow|fly)$.  

Note that $(hollow|fly)$ actually does not give any new information about dogs.  This revision should not change the relative ordering of any worlds with a finite strength of belief.  However, it does result in a change of belief.  
If one is later convinced of the existence of flying dogs, then the fact about hollow bones should be incorporated.
\end{example}

We refer to the reasoning in the preceding example as {\em nearly-counterfactual} revision.  It is essentially a form of counterfactual reasoning, in which hypothetical worlds are considered in isolation.  At the same time, however, we keep a form of conditional memory at higher ordinal levels.  This is not only useful for perspective altering revelations, but we argue it can also be useful for analogical reasoning.  

One important feature that is typically taken as a requirement for conditional reasoning is the Ramsey Test.  In the context of revision by conditional statements, Kern-Isberner formulates the Ramsey Test as follows: 
when revising by a conditional, one would like to ensure that revision by $(\psi|\phi)$ followed by a revision by $\phi$ should guarantee belief in $\psi$ \cite{Kern-Isberner99}.  We suggest that this formulation needs to be refined in order to be used in the case where infinite ranks are possible.

In the case of the flying dog, one is quite likely to accept the conditional $(hollow | fly)$ based on a single report with finite strength.  However, a single report of $fly$ with finite strength will not be believed.  If the antecedent of the conditional is ``very hard'' to believe, then we should not expect the Ramsey Test to hold without some additional condition on the strength of the subsequent report.  The problem, in a sense, is that the notion of believing a conditional is quite different than the notion of believing a fact.  In order to believe $(hollow|fly)$, we simply need to keep some kind of record of this fact for the unlikely case where we discover that flying dogs happen to exist.  
On the other hand, in order to believe $fly$, we really need to make a significant change in our current world view.

\subsection{Levels of Implausibility}
Approaches to counterfactual reasoning are typically inspired to some degree by Lewis, who indicates that the truth of a counterfactual sentence is determined by its
truth in alternative worlds \cite{Lewis73}.  We can represent this idea with $\omega^2$-CFs.  At each limit ordinal $\omega\cdot k$, we essentially have an entirely new plausibility ordering.  As $k$ increases, each such ordering represents an increasingly implausible world.  However, a sufficiently strong observation can force our beliefs to jump to any of these unlikely worlds.  As such, these are not truly counterfactual worlds, because we admit the possibility that they may eventually be believed.

The important property that we can capture with $\omega^2$-CFs is the following:  there are some formulas that may
be true, yet we can not be convinced to believe them based on any finite number of pieces of ``weak evidence.''  This allows us to give the following formal definition of the term {\em nearly counterfactual}.
\begin{definition}
Let $r$ be an OCF.  A formula $\phi$ is nearly counterfactual with respect to $r$ just in case there is no
$\omega$-CF $r^{\prime}$ such that $Bel(r*r^{\prime})\models\phi$. 
\end{definition}
The following is an immediate consequence of this definition.
\begin{proposition}
If $\phi$ is nearly counterfactual with respect to $r$, then there is no finite sequence $r_1,\dots,r_n$ of $\omega$-CFs such that $Bel(r*r_1*\cdots*r_n)\models\phi$.
\end{proposition}
We introduce some useful notation.
\begin{definition}
Let $\phi$ be a formula.  An OCF $r$ is a $\phi$-strengthening iff $Bel(r)=\{s\mid s\models\phi\}$.
\end{definition} 
So, a $\phi$-strengthening is just a ranking function where the minimal states are exactly the models of $\phi$.    For any formula $\phi$, let $(\phi,n)$ be the $\phi$-strengthening of $\phi$ where models of $\phi$ have plausibility $0$ and every other state has plausibility $n$.

\begin{definition}
Let $r$ be an $\omega^2$-CF.  For any limit ordinal $\omega \cdot k$, let 
$r_k$ be the following partial function:
\[
    r_k(s)= \left\{
                    \begin{array}{ll}
                           r(s) \mbox{, if } r(s)=\omega\cdot k + c \mbox{ for some } c\\
			\mbox{undefined otherwise}\\
                    \end{array}
                    \right.\\
\]
\end{definition}
Hence, $r_k$ is just the restriction of $r$ to those states with plausibility values at level $k$.  We say that $\phi$ is {\em believed} at level $k$ if $\{s \mid s\in \min(r_k)\}\models \phi$.  Let $poss(\phi)$ denote the set of natural numbers $k$ such that $s\models \phi$ for some $s$ in the domain of $r_k$.

We can now introduce a form of strengthening with nearly counterfactual conditionals.  In the definition, given an $\omega^2$-CF $r$, we let $deg(s)$ denote the value $k$ such that 
$r(s)=\omega\cdot k +c$.  
\begin{definition}
Let $r$ be an $\omega^2$-CF and let $\psi,\phi$ be formulas where $\phi$ is nearly counterfactual with respect to $r$.  Let $n\in\omega$.
\[
    r*(n,\psi|\phi)(s) = \left\{
                    \begin{array}{ll}
                           r(s) \mbox{, if } deg(s)\not\in poss(\phi)\\
			r*(\psi,n)(s) \mbox{ otherwise}\\
                    \end{array}
                    \right.\\
\]
\end{definition}
We call this function the $n$-stengthening of $\psi$ conditioned on $\phi$.
This function finds all levels of $r$ where $\phi$ is possible, and then strengthens $\psi$ at only those levels.  

\begin{example}
Let $r$ again be the plausibility function
\[
    r(s) = \left\{
                    \begin{array}{ll}
                           \omega \mbox{ if } s\models fly\\
			10\mbox{ if } s\models heavy\wedge\neg fly\\
                           0 \mbox{ otherwise }
                    \end{array}
                    \right.\\
\] 
It is easy to verify
that $fly$ is nearly counterfactual with respect to $r$.  Now suppose that we extend the vocabulary to include 
the predicate symbol $hollow$.  Define a new function $r^{\prime}$ as follows:
\[
    r^{\prime}(s) = \left\{
                    \begin{array}{ll}
                           r(s) \mbox{, if } s\not\models hollow\\
			r(s)+1 \mbox{, if } s\models hollow\\
                    \end{array}
                    \right.\\
\]
This just says that we initally believe our dog does not have hollow bones; however, it is not particularly implausible.  It follows that:
\begin{itemize}
\item $r^{\prime}(s)=\omega$ if $s\models fly\wedge \neg hollow$.
\item $r^{\prime}(s)=\omega+1$ if $s\models fly\wedge hollow$.
\end{itemize}
From these results, it follows that:
\begin{itemize}
\item $r^{\prime}*(2,hollow|fly)(s) = r^{\prime}(s)$, if $s\not\models fly$.
\item $r^{\prime}*(2,hollow|fly)(s) = \omega$, if $s\models fly\wedge hollow$.
\item $r^{\prime}*(2,hollow|fly)(s) = \omega+1$, $s\models fly\wedge \neg hollow$.
\end{itemize}
So, roughly speaking, after strengthening by $(hollow|fly)$, we now believe that hollow bones are more plausible in all hypothetical situations where we believe flying dogs are possible.  
\end{example}

Note that plausibility of a state is only changed at levels where $\phi$ is considered possible.  Since the definition is only applied to nearly counterfactual conditions, this means that only hypothetical states are affected by the strengthening.

It remains to move from conditional strengthening to conditional revision.  Recall that, for any $\omega^2$-CF with $\min(r)=\omega\cdot k + c$, we write $fin(r)=c$.
\begin{definition}
Let $r$ be an $\omega^2$-CF and let $\psi,\phi$ be formulas where $\phi$ is nearly counterfactual with respect to $r$.  
\[
    r*(\psi|\phi)(s) = \left\{
                    \begin{array}{ll}
                           r(s) \mbox{, if } deg(s)\not\in poss(\phi)\\
			r*(\psi,fin(r_k))(s) \mbox{ if } r(s)=\omega \cdot k + c
                    \end{array}
                    \right.\\
\]
\end{definition}
Hence, for revision, we strengthen belief in $\psi$ by the least value that will ensure $\psi$ is believed at level $k$.

Under this definition, we satisfy a modified form of the Ramsey Test.
\begin{proposition}
Let $r$ be an $\omega^2$-CF and let $s$ be a state with $r(s)=\omega\cdot k + c$.  If $r^{\prime}$ is an $\omega^2$-CF with degree of strength larger than $k$ and $Bel(r^{\prime})\models \phi$, then 
$$Bel((r * (\psi|\phi)) * r^{\prime})\models \psi.$$
\end{proposition}
Hence, if we revise by $(\psi|\phi)$ followed by an OCF with ``sufficiently strong'' belief in $\phi$, then $\psi$ will be believed.

\section{Relation to Existing Work}
\subsection{Infinite Plausibility Values}
There has been related work on the use of infinite valued ordinals in OCFs.  In particular, Konieczny defines
the notion of a {\em level} of belief explicitly in terms of limit ordinals\cite{Konieczny09}.  In this work, different ``levels'' are used to represent beliefs that 
are independent in a precise sense.  The lowest level is used for representing an agents actual beliefs about the world, whereas higher levels are used to represent integrity constraints.  Our approach here is different in that we explicitly use the ordering on limit ordinals to represent infinite leaps in plausibility.  This work is also distinguished by the fact that we use ordinal arithmetic on a small class of ordinals to define a simple algebra of belief change.

\subsection{Belief Improvement}
The success postulate ($K*\phi\vdash\phi$) of the AGM framework is clearly incorrect in cases where evidence
is additive.  That is to say, there are situations where a single observation is not sufficient to convince an agent
to believe a particular fact.  {\em Improvement operators} \cite{Konieczny08} are belief change operators that address
this issue by introducing a new set of postulates.  The most important postulate states that an improvement 
operator $\circ$ must have the property that:
\begin{itemize}
\item[{\bf (I1)}] {\em There exists $n\in\mathbf{N}$ such that } $B(\Psi\circ^n\phi)\vdash\phi$.
\end{itemize}
Here $\Psi$ is an epistemic state, and $B(\cdot)$ maps an epistemic state to the minimal elements of the underlying ordering.  Hence, an improvement operator has the property that an agent will be convinced to believe $\phi$ after a finite number of improvements.  The remaining postulates for a {\em weak improvement operator} are essentially
the DP postulates applied to the operation $\circ^n$ obtained from {\bf (I1)}.  We refer the reader to  \cite{Konieczny08} for the complete list of postulates.

We define an analog of  {\bf (I1)} as follows.  If $r_\phi$ denotes a $\phi$-strengthening, we can express 
the condition as follows.
\begin{itemize}
\item[(I$^*$)] {\em There exists $n\in\mathbf{N}$ such that } $Bel(r*^nr_\phi)\models\phi$.
\end{itemize}
The truth of this property depends on the degrees of strength of the functions. 

\begin{proposition}\label{rop}
If $r$ is an $\omega$-CF and $r_\phi$ is a $\phi$-strengthening with finite strength, then {\bf I$^*$} holds.
\end{proposition}
For an epistemic state $\Psi$ defined by $\prec_\Psi$, let $r_\Psi$ be the {\em canonical representation} of $\Psi$.\footnote{If $s$ is in the $n^{th}$ level of $\Psi$ , then $r_\Psi(s)=n$}  Define $\circ_n$ such that $\Psi\circ\phi$ is obtained by taking the ordering induced by $r_\Psi*r(\phi,n)$.
\begin{proposition}\label{pop}
For any $n\in\mathbf{N}$, the operator $\circ_n$ is a weak improvement operator.
\end{proposition}
We call $\circ_n$ a {\em finite improvement operator}, because the degree of strength is finite.  This result is essentially a corollary of Proposition \ref{rop}, and it suggests that our $*$ operation based on normalized addition is actually the natural extension of {\em improvement} to the setting of $\omega$-CFs.  

The advantage of infinite plausibility values is that they give us greater flexibility in modelling improvement.
\begin{proposition}
If $r$ is an $\omega^2$-CF and $r_\phi$ is a $\phi$-strengthening with finite strength, then {\bf I$^*$} does not hold.
\end{proposition}
This result essentially states that ({\bf I1}) is not a sound property for $*$ if we allow infinite plausibility values.  This distinction can be seen in our running example.  There is no finite number of improvements
that will force the agent to believe that dogs can fly.

It is actually difficult to express the analog of Proposition \ref{pop} in the context of $\omega^2$-CFs, because a total pre-order over states can not capture the ``infinite jumps'' in plausibility encoded by $\omega^2$ ordinal ranks.  But it is possible to define a correspondence between sequences of orderings and ordinals in $\omega^2$. 

\begin{definition}
Let $r$ be an $\omega^2$-CF where $\max(r)=\omega \cdot d + b$ for some $d, b$.  For $i\le d$, let $r_i$ denote the function defined as follows:
\begin{enumerate}
\item $domain(r_i)=\{s \mid r(s)=\omega\cdot i + c\}$.
\item If $r(s)=\omega\cdot i + c$, then $r_i(s) = c$.
\end{enumerate}
\end{definition}
The following propositions are immediate.
\begin{proposition}
Each $r_i$ is a $\omega$ ranking, and there exists an $\omega$-CF such that $r_i^{\prime}\sim r_i$
\end{proposition}
\begin{proposition}
For any $\omega^2$-CF $r$ over a vocabulary ${\mathbf P}$ with $deg(r)=d$, there is an extended vocabulary ${\mathbf P}_1$ and a sequence $r_0,\dots,r_d$ of $\omega$-CFs such that, for each $i\le d$, the $r_i$ is equivalent (i.e. $\sim$) to the restriction of $r$ to ordinals of degree $i$.
\end{proposition}
This result is proved by just extending the vocabulary appropriately with propositional variables that make each infinite jump in the ordinal value definable.  By breaking $r$ into a set of $\omega$-CFs, it follows that 
{\bf (I$^*$)} holds at level $d$ when $r_\phi$ has degree of strength $\omega\cdot d$.  Therefore, belief change by normalized addition on $\omega^2$-CFs can really just be seen as a finite collection of improvements as each level.
The important point, however, is that no finite sequence of improvements at level $d$ will ever impact the actual beliefs at lower levels.

\subsection{Conditional Belief Revision}
Conditional belief revision was previously addressed by Kern-Isberner, who proposes a set of rationality postulates for conditional revision \cite{Kern-Isberner99}.  A concrete approach to conditional revision is also proposed, through
the following $\omega$-CF :
\[
    r*(\psi|\phi)(s) = \left\{
                    \begin{array}{ll}
                           r(s) - r(\psi|\phi) \mbox{, if } s\models \phi\wedge \psi\\
			r(s) + \alpha + 1 \mbox{, if } s\models \phi\wedge\neg \psi\\
			r(s) \mbox{, if } s\models\neg \phi\\
                    \end{array}
                    \right.\\
\]
where $\alpha=-1$ if $r(\{\phi,\psi\}) < r(\{\phi\})$, and $\alpha=0$ otherwise.  This operation satisfies all of the postulates for conditional revision, as well as the Ramsey Test.  We remark, however, that this approach is not well-defined if we allow infinite plausibility values because of the ordinal subtraction on the right hand side.  We suggest that this is not just a formal artefact of the theory; conditionals that are ''almost certainly'' false actually must be treated slightly differently.

In our approach, we essentially require the evidence for $\phi$ to be substantially stronger than the evidence for the conditional.  We suggest that our beliefs following conditional revision should be changed in sort of an infinitesimally small way.  While our beliefs about the actual world do not change, our beliefs about some (nearly) impossible world do, in fact, change.

Note that it is actually possible to reconcile our approach with Kern-Isberner's approach, by using the conditional revision above on each level $r_k$ of the initial OCF $r$.  At present, we are using a simple strengthening on each level, which actually flattens the plausibility structure after ordinal addition.  A combined approach could respect the infinite jumps in plausibility, while satisfying the postulates for conditional revision at each level.  We leave an investigation of this combined approach for future work.

\section{Discussion}
\subsection{Conclusion}
In this paper, we have explored the use of infinite ordinals for reasoning about belief change and conditional reasoning.  We have shown that allowing plausibility values to range over $\omega^2$ results in a belief algebra that is only slightly more complicated, and we gain an expressive advantage.  In particular, we can represent situations where stubbornly held beliefs are resistant to evidence to the contrary.  We have demonstrated that this results in a slightly more expressive class of improvement operators where evidence increases relative belief, but no finite number of improvements will actually lead to a change in the belief state.  Finally, we addressed so-called ``nearly counterfactual'' revision, where we incorporate information that is conditional on a highly unlikely statement.

\subsection{Future Work}
This paper is a preliminary exploration into different applications and formal properties of infinite valued ordinal conditional functions.  It remains to move beyond $\omega^2$-CFs, to completely characterize the relationship with improvement operators, and to consider further practical applications.  

In the present framework, we have discussed nearly counterfactual reasoning as a tool for keeping a sort of ``memory'' about unlikely situations, in order to incorporate this information later if necessary.  But there is also a natural kind of reasoning that would allow us to use conditionals to reason by analogy about the actual state of the world.  Consider the following well-known ambiguity from 
\cite{Lewis73}, and originally attributed to Quine:
\begin{enumerate}
\item If Caesar was president, he would use nuclear weapons.
\item If Caesar was president, he would use catapults.
\end{enumerate}
As a conditional, we could write both as $(W|C)$, where $W$ stands for a weapon that would be used and $C$ is the condition ``Caesar is president.''  But (1) suggests that we condition by imagining Caesar alive in the current world.  So this is a conditional statement interpreted in the current state of the world.  On the other hand, (2) suggests that we consider what would happen in some past world where Caesar exists.  

Now suppose that we believe a certain politician is actually very similar to Caesar.  If we believe that Caesar would use nuclear weapons, then we may conclude that this ``real'' politician would also use nuclear weapons.  Formally, we could proceed as follows:  if some hypothetical world is isomorphic to the current state of the world when we restrict the vocabulary (to not include Caesar), then we can use inferences about the hypothetical world to draw conclusions about the actual world.  This is a form of {\em ampliative reasoning} that we intend to explore through $\omega^2$-CFs in future work.


\bibliographystyle{alpha}

\end{document}